\documentclass[10pt,twocolumn,letterpaper]{article}

\usepackage{cvpr}              
\usepackage{multirow}
\usepackage{algorithm}      
\usepackage{algpseudocode}
\usepackage[accsupp]{axessibility}

\definecolor{cvprblue}{rgb}{0.21,0.49,0.74}
\usepackage[pagebackref,breaklinks,colorlinks,allcolors=cvprblue]{hyperref}

\def\paperID{BEAM-12} 
\def\confName{CVPR}
\def\confYear{2025}

\title{Quantum Federated Learning for Multimodal Data: A Modality-Agnostic Approach} 
\def\spaces{~~~~~~}
\author{Atit Pokharel\textsuperscript{\dag}\spaces{}Ratun Rahman\textsuperscript{\dag}\spaces{}Thomas Morris\textsuperscript{\dag}\spaces{}Dinh C. Nguyen\textsuperscript{\dag}\\
\textsuperscript{\dag{}}Department of Electrical and Computer Engineering, The University of Alabama in Huntsville, USA\\
{\tt\small ap1284@uah.edu, rr0110@uah.edu, tommy.morris@uah.edu, dinh.nguyen@uah.edu }
}

\begin{document}
\maketitle
\begin{abstract}
Quantum federated learning (QFL) has been recently introduced to enable a distributed privacy-preserving quantum machine learning (QML) model training across quantum processors (clients). Despite recent research efforts, existing QFL frameworks predominantly focus on unimodal systems, limiting their applicability to real-world tasks that often naturally involve multiple modalities. To fill this significant gap, we present for the first time a novel multimodal approach specifically tailored for the QFL setting with the intermediate fusion using quantum entanglement. Furthermore, to address a major bottleneck in multimodal QFL, where the absence of certain modalities during training can degrade model performance, we introduce a Missing Modality Agnostic (MMA) mechanism that isolates untrained quantum circuits, ensuring stable training without corrupted states. Simulation results demonstrate that the proposed multimodal QFL method with MMA yields an improvement in accuracy of 6.84\% in independent and identically distributed (IID) and 7.25\% in non-IID data distributions compared to the state-of-the-art methods. 
\end{abstract}
    
\section{Introduction}   
\label{sec:intro}
Quantum federated learning (QFL) has recently emerged as an innovative framework that merges two groundbreaking technologies: quantum computing and federated learning (FL)\cite{innan2024fedqnn}. The key idea of classical FL is to allow decentralized clients to train models locally, sharing only model parameters with a central server, thereby preserving data locality and reducing communication overhead \cite{nguyen2021federated, perf2024}. However, the bottleneck of traditional FL is raised by growing data dimensionality and complex optimization requirements \cite{shi2022towards,rahman2025electrical,rahman2024electrical}. Quantum computing introduces the possibilities in large-scale machine learning by bringing in the quantum advantage from its core principles of superposition and entanglement. Quantum-based models often accelerate convergence and improve representational power in diverse machine-learning tasks where classical approaches fail to perform \cite{huang2021power}. In this context, QFL enables the collaboration of quantum machine learning (QML), e.g., a shared quantum neural network (QNN), model training over quantum processors (clients) with an aggregator (quantum server). 

\textcolor{black}{\textbf{Our Key Motivations}: The majority of literature works \cite{chen2021federated, huang2022quantum, qiao2024transitioning, park2024dynamic} focus on single-modal QFL settings, where data across quantum clients is assumed to be of the same type or modality. However, in real-life applications, multimodal QFL, which integrates multiple data modalities (e.g., a combination of images or audio data in emotion recognition), is often necessary to address more complex problems. While single-modal QFL simplifies the learning process by working with uniform data, multimodal QFL enables richer and more comprehensive insights by leveraging diverse data sources. Developing robust QFL systems that can handle multimodal settings is essential for advancing practical quantum machine learning applications.  Another significant challenge is the potential absence of continuous data across all modalities \cite{zhao2024deep}, as some sensors or data sources may become inactive or faulty. If these missing modalities are not properly addressed, they can significantly degrade the overall performance of the system \cite{reza2024robust}. To tackle this issue, we propose a QFL framework specifically designed for multimodal data, incorporating robust mechanisms to effectively handle missing modalities for robust multimodal QFL.}

\textcolor{black}{\textbf{Paper Contributions}:
Motivated by limitations in the literature, we propose mmQFL, a multimodal quantum federated learning framework with cross-modal correlations and a missing modality-agnostic mechanism. }
Our key contributions are summarized as follows
\begin{itemize}
    \item We introduce a multimodal approach in the QFL scenario that integrates multiple modalities for a more robust representation.  \textit{To the best of our knowledge, this work is the first to study multimodal approach, particularly in a QFL setting along with deeper inter-modal correlation and better handling of missing modalities.}
    \item We design a quantum fusion layer with entanglement-based fusion for deeper inter-modal correlations in the quantum state space.  A missing modality agnostic mechanism using no-op gates and context vectors is introduced to isolate incomplete modalities and maintain model stability. 
    \item Extensive simulations with the CMU-MOSEI dataset under independent and identically distributed (IID) and non-IID distributions demonstrate the merits of our approach in terms of better training performance and stability compared with state-of-the-art schemes.
\end{itemize}

\begin{figure*}
    \centering
    \includegraphics[width=0.99\linewidth]{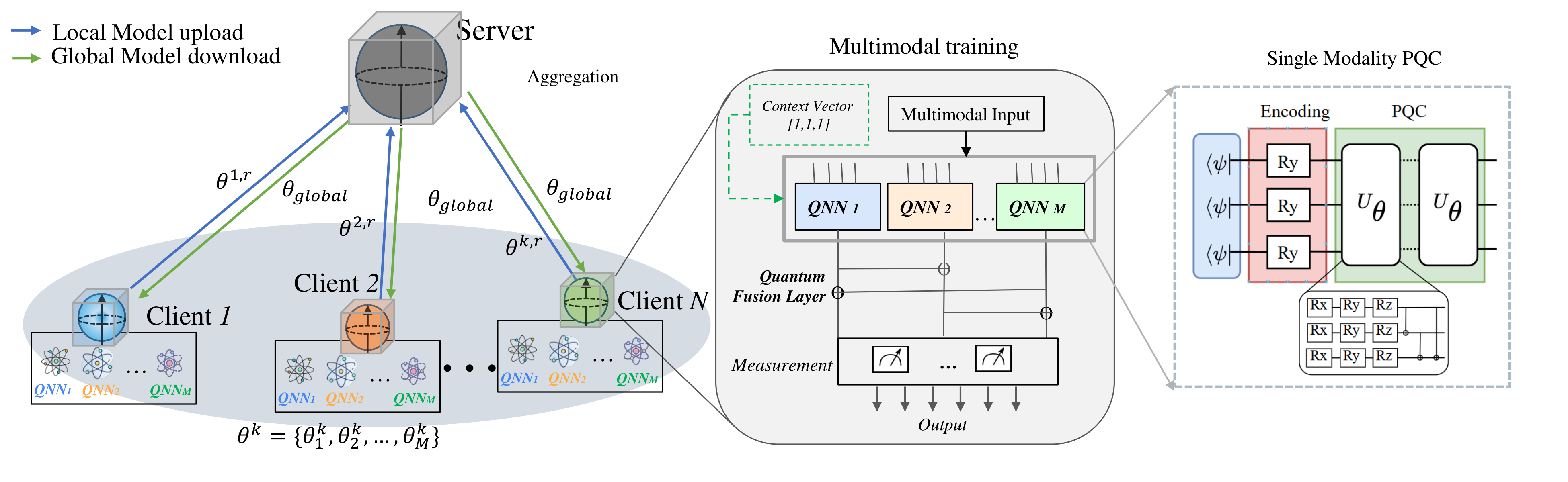}
    \caption{An overview framework of the proposed mmQFL approach, where each quantum client has $M$ modalities. Each client trains separate QNN models for each modality and fuses them using a quantum fusion layer before sending the fused model to the quantum server for model aggregation. }
    \label{fig: overview}
    \vspace{-5mm}
\end{figure*}
\section{Related Works}
\label{sec:formatting}
\subsection{Quantum Machine Learning}
QML combines quantum computing concepts with conventional machine learning approaches, resulting in possible speedups and performance improvements for many computational workloads \cite{biamonte2017quantum}.  Recent research \cite{schuld2019quantum, havlivcek2019supervised} has focused on quantum neural networks (QNNs), quantum support vector machines (QSVMs), and quantum reinforcement learning methods.  QML models frequently use variational quantum circuits that use quantum entanglement and superposition to represent complicated data connections \cite{kahou2016emonets,benedetti2019parameterized}.  Despite the potential, obstacles such as noise in quantum circuits, limited quantum hardware resources, and scalability remain key issues \cite{cerezo2021variational}. However, QML frequently encounters issues linked to the centralization of sensitive data, which might raise privacy concerns and a considerable quantum of resources on the server.

\subsection{Quantum Federated Learning}
QFL blends federated learning (FL), a privacy-preserving decentralized machine learning technique, with quantum computing technologies to improve privacy and learning efficiency \cite{chen2021federated,rahman2024improved}. Classical FL allows numerous clients to train models without exchanging raw data, ensuring data privacy and lowering communication overhead \cite{mcmahan2017communication,feng2023fedmultimodal,uddin2024false}. QFL builds on this paradigm by including quantum algorithms for safe aggregation and privacy improvement, leveraging quantum encryption methods and quantum-enhanced optimization procedures \cite{li2021quantum}. Recent research has investigated quantum-secure multiparty computation for federated learning \cite{huang2022quantum}, quantum-enhanced FL aggregation approaches \cite{zhao2021quantum}, and showed possible gains in convergence speed and security resilience \cite{chehimi2022quantum,qu2023qnmf}.

\subsection{Multimodal Learning}
Unimodal learning \cite{zhang2019face, noroozi2017audio, kratzwald2018deep, pokh2019} struggles to achieve adequate performance due to the lack of perspectives. Multimodal learning integrates and processes various modalities, including text, audio, and visual input, to improve prediction performance and generate strong representations \cite{baltruvsaitis2018multimodal,rahman2024multimodal}.  Recent studies use deep learning approaches, namely transformer structures, to represent complicated interactions across modalities \cite{vaswani2017attention}.  These approaches have proven successful in tasks such as sentiment analysis \cite{zadeh2017tensor}, video classification \cite{carreira2017quo}, and multimodal emotion recognition \cite{poria2017review,rahman2024multimodal}.  However, multimodal learning confronts obstacles such as data heterogeneity, modality alignment, and efficient fusion processes \cite{liang2022foundations}.  Multimodal scenarios in quantum machine learning have been explored in \cite{li2025qmlsc, zheng2024quantum, qu2024qmfnd}, where researchers have developed quantum models that fuse diverse modalities via PQCs. However, no prior works have been thoroughly conducted to investigate multimodal settings in QFL.
\subsection{Missing Modalities in Multimodal Learning}
Missing or incomplete modalities are a common issue in multimodal learning, affecting model performance and generalization capabilities \cite{ma2021smil}.  Existing research in classical multimodal learning has paved the way to address this issue of modality imputation, dropout training, and cross-modal generative techniques \cite{pham2019found}.  Autoencoders and generative adversarial networks (GANs) are used for imputing missing modalities, allowing models to acquire robust representations with incomplete inputs \cite{tran2017missing}. However, the handling of the missing modality problem in multimodal QFL is a significant challenge due to the different nature of quantum machine learning.



\section{Proposed Multimodal QFL(mmQFL) Framework}

\subsection{Overview of the Framework}

\textcolor{black}{
The mmQFL framework consists of a network of quantum processors (clients) denoted as the set $\mathcal{N}$ that collaborate to train a shared multimodal QML model with a quantum server. The complete training framework is structured as follows
\begin{itemize}
    \item Each client’s dataset contains multiple modalities with dedicated QNN models to ensure modality-specific learning. Each modality features are encoded into the quantum states, which are to be processed independently through parameterized quantum circuits (PQCs), preserving unique modality characteristics.
    \item An intermediate quantum fusion layer with full entanglement is introduced before measurement, operating between PQCs of all modalities. This layer also uses trainable parameters to establish cross-modal correlations directly in the quantum state space. The final measurement produces classical outputs, which are used to generate the classification result and calculate loss value.
    \item Missing modalities are addressed using a context vector that tracks the modality data availability. During a missing data modality case, this forces qubits into the zero states. The untrained QNN block is isolated during fusion and prevents corrupted states from degrading the performance.  This technique has been discussed in classical machine learning literature \cite{wang2023multi}, and we extend it to the QML scenario through our mmQFL framework.
    \item After local training, clients upload their modality-specific and fused models to the quantum server for weighted aggregation. The updated global parameters are then shared with clients, completing the mmQFL cycle. The evaluation uses multi-modal data points to produce a single classification output.
\end{itemize}}


\subsection{Problem Setting}
Each client denoted by $k$ contains a local dataset $D_k$ having $M$ modalities. A data sample $x^k$ with label $y^k$ on dataset $D_k$ is described as
 \begin{equation}\label{eq:multimodal-input} 
\mathbf{x}^k = \bigl( \mathbf{x}^{k}_1, \;\mathbf{x}^{k}_2, \dots, \mathbf{x}^{k}_M \bigr),
\end{equation}
where $M$ is the total number of available modalities, and each $\mathbf{x}_m^k \in \mathbb{R}^{d_m}$ is a feature vector corresponding to modality $m$. Each quantum client has a modality model $\boldsymbol{\theta_m^k}$ for every modality dataset $D_m^k$ obtained by training its tailored PQC designed to capture intricate relations between input and its label. Thus, a client incorporates a set of models $\boldsymbol{\theta^k}$, which is formulated and discussed in \ref{subsection: fusion}.

\subsection{Encoding Multimodal Data into Quantum States and Training}
The foremost step in a QNN involves encoding the classical feature vector into quantum states. Here, the dimensionality-adjusted feature vector of each modality can be \emph{amplitude-encoded} into a normalized quantum state \cite{ranga2024quantum}. Each modality $m$ is encoded by a separate amplitude-encoding circuit, to process them further in $M$ distinct PQCs. For modality $m$ on client $D_k$, we have an $n_m$-qubit encoder mapping as
\begin{equation}
    \label{eq:modality-encoding}
    \mathbf{x}_m^k \;\mapsto\; \lvert{\psi_{m,\text{enc}}^k}\rangle \;=\; \sum_{i=1}^{N_m} \bigl( x^k_{m,i} \bigr) \lvert{i}_m\rangle,
\end{equation}
where $\|\mathbf{x}_m^k\| = 1$, and $N_m = 2^{n_m}$ and \(\lvert \psi_{m,\text{enc}}^k \rangle\) represents the quantum state vector in a Hilbert space resulting from the encoding.
The encoded state is then processed by an \(L\)-layer parameterized quantum circuit (PQC), where each layer \(l\) (for \( l = 1, 2, \dots, L \)) sequentially applies a trainable unitary transformation, allowing the quantum state to evolve as
\begin{equation} \label{eq: pqc transformation}
\lvert \psi_{m,l}^k \rangle = U_{m,l}^k(\boldsymbol{\phi}_{m,l}^k)\,\lvert \psi_{m,l-1}^k \rangle,\quad \text{with}\quad\lvert \psi_{m,0}^k \rangle = \lvert{\psi_{m,\text{enc}}^k}\rangle, 
\end{equation}
where, ${\phi}_{m,l}^k$ is the set of rotational angles in single qubit rotational gates ($R_x$, $R_y$, and $R_z$) of layer $l$, which acts as the trainable model parameter in PQC of modality $m$.

Each layer’s unitary \( U_{m,l}^k(\boldsymbol{\phi}_{m,l}^k) \) comprises two sequential operations:

\begin{itemize}
    \item {Single-qubit rotation gates} \(U^{(l)}_{\text{rot}}\): Applies rotation gates parameterized by \(\boldsymbol{\phi}_{m,l}^k\) to each qubit individually.
    \item {Entangling layer} \(U^{(l)}_{\text{ent}}\): Establishes correlations between qubits through a cyclic sequence of controlled-NOT (CNOT) gates.
\end{itemize}
Thus, the complete sequential unitary transformation for layer \(l\) is given by
$
U_{m,l}^k(\boldsymbol{\phi}_{m,l}^k) = U^{(l)}_{\text{ent}}\,U^{(l)}_{\text{rot}}.
$
The overall sequential transformation of the PQC across all layers is represented as
\begin{equation}\label{eq: unitary transformation in all layers}
U_{m,\text{total}}^k(\boldsymbol{\theta}_m^k) = \prod_{l=1}^{L} U_{m,l}^k(\boldsymbol{\phi}_{m,l}^k),
\end{equation}
where \( \boldsymbol{\theta}_m^k = (\boldsymbol{\phi}_{m,1}^k, \dots, \boldsymbol{\phi}_{m,L}^k) \). The final output quantum state after these sequential transformations evolves as
\begin{equation} \label{eq: quantum state after pqc in m}
\lvert{\psi_{m,\text{out}}^k(\boldsymbol{\theta}_m^k)}\rangle = U_{m,\text{total}}^k(\boldsymbol{\theta}_m^k)\,\lvert{\psi_{m,\text{enc}}^k}\rangle. 
\end{equation}
Here, $\boldsymbol{\theta}_m^k$ represents the set of all trainable parameters for the entire modality $m$ on client $k$.

\subsection{Fusion of Modalities and Loss Function}\label{subsection: fusion}
To combine these modality-specific quantum states into a unified representation, we incorporate an additional \emph{quantum fusion layer} that entangles the separate modality PQCs into a comprehensive modal inspired by the intermediate fusion in the classical approach \cite{boulahia2021early}. This step is only involved in the final framework and not included in the initial study (Table \ref{tab:mm}) where the contribution of individual modality is studied. During this, the evolved state after Eq. \eqref{eq: quantum state after pqc in m} is further processed to the measurement layer where the Pauli-Z observable for each qubit is observed with a similar operation as in Eq. \eqref{eq:modality-obs}. The only difference is the absence of a quantum fusion layer before the measurement, due to which the set of models $\boldsymbol{\theta^k}$ becomes
$
  \bigl\{
    \boldsymbol{\theta}_1^k,\boldsymbol{\theta}_2^k,\dots,\boldsymbol{\theta}_M^k
  \bigr\}.
$
This concatenation is analogous to the late-fusion approach, where the modalities are fused after the training. In this setup, the next step involves federation, as discussed in \ref{subsection: FL}. After the quantum server distributes the global models for all modalities, each modality is evaluated individually to assess its contribution.

Whereas, in the final model with a quantum fusion layer, information exchange between modalities is developed so that the final quantum state representation involves cross-modal correlations.
The total system comprises $n_q = \sum_{m=1}^M n_m$ qubits (one QNN model per modality) and after each modality’s local unitary $U_m^k(\boldsymbol{\theta}_m^k)$, we apply an entangled inter-modality circuit $U_{\text{fusion}}^k(\boldsymbol{\theta}_{\text{fusion}}^k)$ acting on \emph{all} $n_q$ qubits given by
\begin{equation}\label{eq: fusion}
    \lvert{\Psi_{\text{out}}^k(\boldsymbol{\theta}^k)}\rangle
    \;=\;
    U_{\text{fusion}}^k\!\bigl(\boldsymbol{\theta}_{\text{fusion}}^k\bigr)
    \bigotimes_{m=1}^{M}
    \lvert{\psi_{m,\text{out}}^k(\boldsymbol{\theta}_m^k)}\rangle, 
\end{equation}
where 
\(
  \boldsymbol{\theta}^k 
  \;=\;
  \bigl\{
    \boldsymbol{\theta}_1^k,\dots,\boldsymbol{\theta}_M^k,\;\boldsymbol{\theta}_{\text{fusion}}^k
  \bigr\}
\)
is the concatenated form of models of \emph{all} modalities on client $k$. Here $\boldsymbol\theta^k_{\text{fusion}}$ is the set of trainable parameters in an additional layer in the fusion layer. By stacking this layer, the fusion mechanism improves multimodal correlation while holding the flexibility to compensate for missing modalities.

\subsection{Measurement and Loss Function}
A final measurement operator $O$ (or set of Pauli operators on the $n_q$ qubits) produces classical outputs. The measurement retrieves the Pauli-Z observable in classical values  as
\begin{equation}\label{eq:modality-obs}
p^k(\boldsymbol{\theta}^k)
\;=\;
\bigl\langle
\psi_{\text{out}}^k(\boldsymbol{\theta}^k)
\big|
\,O_m
\big|
\psi_{\text{out}}^k(\boldsymbol{\theta}^k)
\bigr\rangle,
\end{equation}
where $O_m$ is a chosen Pauli-Z measurement operator.
The measured qubit states are used to produce a predicted label~$\hat{y}^k$. Denoting the measurement outcome distribution by $p(\hat{y}^k \mid \boldsymbol{\theta}^k)$, the local loss on client $k$ with dataset $D_k$ is 
\begin{equation} \label{eq:loss function}
L_k(\boldsymbol{\theta}^k) 
\;=\;
\frac{1}{D_k}
\sum_{x=1}^{D_k}
\ell\Bigl(
  p\bigl(\hat{y}^k \mid \boldsymbol{\theta}^k,\;\mathbf{x}^k_x\bigr),
  \;y^k_x
\Bigr),
\end{equation}

where $\ell$ is the per-sample loss value. Note that the loss calculation for the individual modality evaluation incorporates its respective modality dataset $D_m^k$. The optimization in the PQC is performed using a classical optimizer, with gradients estimated via the parameter-shift rule \cite{wierichs2022general}, ensuring efficient computation of updates for the trainable parameters in the quantum circuit.

\subsection{Missing Modality Agnostic (MMA) Approach}
In a scenario where $M'$ out of $M$ modalities are missing on client dataset $D_k$, certain encoded states or sub-circuits are disabled. Formally, let $c \in \{0,1\}^{M}$ be a \emph{context vector}, where $c_m = 0$ indicates modality $m$ is missing. Then, in the encoding step, we replace 
\(\mathbf{x}_m^k\mapsto \lvert{\psi_{m,\text{enc}}^k}\rangle\)\
with a blank (no-op gate) if $c_m=0$. 

The fusion circuit $U_{\text{fusion}}^k$ acts on QNN models from all modalities. When a modality is missing, the corresponding qubits remain initialized in a fixed zero state using no-op gates, preventing unwanted interference.  This \emph{missing-modality-agnostic} design ensures that the QNN remains functional even under partial modality availability. Without the MMA mechanism, the PQC of the missing modality becomes randomly initialized or receives garbage values from faulty sensors, leading to corrupted quantum states. These inaccurate states get fused with other properly trained modalities, degrading overall training performance and classification accuracy. In contrast, when MMA is employed, the context vector immediately flags missing modality data. Consequently, missing modalities remain isolated in the zero state via no-op gates, preventing interference and significantly stabilizing the training process. Although the potential information loss due to the absence of modality isn't recovered, the MMA scheme prevents negative impact due to the corrupted states.

\subsection{Incorporating Federated Learning}\label{subsection: FL}

The QFL algorithm is a decentralized training framework similar to classical federated learning. The process is iterative and progresses in \emph{communication rounds} indexed by \(r\). Each round involves three key steps as follows

\begin{enumerate}
    \item \textbf{Client Initialization and Local Update:}  
    At the start of round \(r\), the server broadcasts the current global model 
    \(\boldsymbol{\theta^{(r)}_{\text{global}}}\) 
    to all clients. Each client initializes its local model as 
    \(\boldsymbol{\theta}^{k, r-1} \leftarrow \boldsymbol{\theta^{(r)_{\text{global}}}}\)  
    and trains it on its local dataset using gradient-based updates to minimize its loss function given by Eq. \eqref{eq:loss function}. 
    After a predefined number of local iterations, each client obtains its trained parameter vector in global round $r$ as
    \(\boldsymbol{\theta}_k^{(r)}\).

    \item \textbf{Aggregation:}  
    After the local training in the global round \(r\), the server collects model parameters from all clients. The total data size is \( D = \sum_{k=1}^{\mathcal{N}} D_k \). Each client maintains parameters \(\theta_{m}^{k,r}\) for each modality and \(\theta_{\text{fusion}}^{k, r}\) for the fusion model. The global update for both modality-specific and fusion parameters is expressed as
\begin{equation}\label{eq:unified-update}
\theta_{global}^{(r+1)} =
\frac{1}{D}
\sum_{k=1}^{\mathcal{N}}
 D_k\,\theta_{m}^{k, r},
\quad
\forall m \in \{1,\dots,M, \text{fusion} \}.
\end{equation}

The aggregated global model $\theta_{\text{global}}^{(r+1)}$ is then represented as
$
\Bigl\{
  \theta_{1}^{(r+1)},\,
  \theta_{2}^{(r+1)},\dots,
  \theta_{M}^{(r+1)},\,
  \theta_{\text{fusion}}^{(r+1)}
\Bigr\}.
$
The weighted averaging ensures that each client's contribution is proportional to its dataset size, improving robustness in non-IID data settings.

\end{enumerate}
This iterative process continues until the global model converges to an optimal solution, producing the final trained model 
\(\boldsymbol{\theta_{\text{global}}}^{(R)}\), 
where $R$ is the total number of global communication rounds.
  
\subsection{Algorithm}
Algorithm \ref{alg:MQFL} summarizes the workflow of mmQFL with missing modality agnostic, where the core training process starts with the initialization of the global model (line 2), which is then distributed to all the clients (lines 4-7). The client trains the QFL model with multimodal data, separately for each modality. The presence or absence of a certain modality is considered via a context vector and a fusion mechanism is performed (lines 9-10). The loss function is then calculated and all model parameters are updated iteratively during local iterations (lines 11-13). The locally trained models in the global epoch $r$ are then sent to the server from all the clients for aggregation. Finally, the server aggregates the received local models and broadcasts the new global model again. This whole process repeats until convergence is reached. 
\begin{algorithm}[ht]
\footnotesize
\caption{Multimodal Quantum Federated Learning}
\label{alg:MQFL}
\begin{algorithmic}[1]
    \State \textbf{Input:} Global rounds $R$, clients $K$, local multimodal datasets $\{\mathcal{D}_k\}_{k=1}^K$.
    \State \textbf{Output:} Final global model $\boldsymbol{\theta}_{global}^{(R)}$.
    \State \textbf{Initialize:} $\boldsymbol{\theta}_{global}^{(0)}$.
    \For{$r = 0$ and round $r$ to $R-1$}
        \State Server broadcasts $\boldsymbol{\theta}_{global}^{(r-1)}$.
        \For{each client $k$ in parallel}
            \State Set $\boldsymbol{\theta}^{k,r} \gets \boldsymbol{\theta}_{global}^{(r)}$.
            \For{each local epoch $t$ and sample $\mathbf{x}^k \in \mathcal{D}_k$}
                \State For each modality $m$: \textbf{if} available, encode \Statex \hspace{5.5em}$\mathbf{x}_m^k$ via Eq.~\eqref{eq:modality-encoding} and apply $U_m^k(\boldsymbol{\theta}_m^k)$ 
                \Statex \hspace{5.5em}(Eq.~\eqref{eq: quantum state after pqc in m}); \textbf{else}, use a no-op.
          
                \State Fuse modality outputs using Eq.~\eqref{eq: fusion} and \Statex \hspace{5.5em}measure as in Eq.~\eqref{eq:modality-obs} 
                to obtain 
                \Statex \hspace{5.5em}prediction $\hat{y}^k$.
                \State Compute loss $\ell\Bigl(p(\hat{y}^k\mid\boldsymbol{\theta}^k,\mathbf{x}^k),y^k\Bigr)$.
                \State Update the optimized parameters $\boldsymbol{\theta}_k$.
            \EndFor
            \State Client sends $\boldsymbol{\theta}^{k,r}$ to server.
        \EndFor
        \State Server aggregates using Eq.~\eqref{eq:unified-update}.
    \EndFor
    \State \textbf{Return:} $\boldsymbol{\theta}_{global}^{(R)}$.
\end{algorithmic}
\end{algorithm}
\section{Experiments}
\subsection{Dataset}
For our proposed method, we have used the Carnegie Mellon University Multimodal Opinion Sentiment and Emotion Intensity (CMU-MOSEI) dataset \cite{zadeh2018multimodal}. The CMU-MOSEI dataset is a large multimodal dataset that is frequently utilized in the scientific community for sentiment analysis, emotion recognition, and multimodal language processing.  Extracted from YouTube videos, it contains text, audio, and video data from over 23,500 sentence utterances from over 1,000 distinct speakers covering more than 250 themes.  Sentiment on a continuous scale ranging from -3 (very negative) to +3 (extremely positive) and six fundamental emotions—happiness, sorrow, anger, fear, disgust, and surprise—on a scale ranging from 0 to 3 have been assigned for each statement in the dataset. 

\textbf{Data Processing} In this research, we use the audio, image, and text modalities from the CMU MOSEI dataset. We decide to extract 2 frames per second from the video frames to capture the image data. Then we resize the image to a smaller size of $32\times 32$ pixels to facilitate quantum encoding. For the audio, the corresponding audio segment is sampled at 20 Hz and it gets converted to Mel-frequency cepstral coefficients (MFCCs) format as this format is more manageable for quantum algorithms \cite{abdul2022mel}. This extracts features such as voice intensity and frequency and efficiently reduces the feature dimension. For the text data, we utilize the pre-extracted embeddings. The dimensionality of these features is adjusted using a dense layer before applying quantum encoding. This dataset is inherently synchronized across modalities, ensuring that text, audio, and image data are aligned which ensures temporal coherence without requiring additional synchronizations. 

\subsection{Simulation Settings}
We present a QFL system that trains QNN data processing using ten noisy intermediate-scale quantum (NISQ) devices for the local dataset of each modality and a single quantum server. Given extensive research on quantum noise mitigation \cite{hama2024quantum,singh2023mid,takagi2022fundamental, sharma2022implementation} and the advancement of quantum computers toward less noisy environments, we focus on minimal noise, as noisy simulations are not the primary scope of our research. Thus, we use quantum depolarizing noise with a depolarizing probability ranging from 0.001 to 0.05. The quantum simulation setup is supported by the state vector simulator \textit{torchquantum}. For the QNN model, we have named iQNN for the image-classification quantum model, aQNN for the audio-classification quantum model, and tQNN for the text-classification quantum model. 

\textbf{Non-IID data distribution.} In our FL study, we investigate non-IID data distributions across ten clients for images and five clients for audio, which included 6000 unlabeled training pictures saved on the server. The clients have random rata data distributions with different feature spaces.
The number of features for the emotions happiness, sadness, anger, disgust, surprise, and fear is 12500, 6000, 5000, 4050, 2250, and 1900 respectively. In our distribution, the input data is already heterogeneous as the number of data points on each client varies by a significant number. This heterogeneity is further propagated into the quantum domain through the implementation of angle encoding, which directly translates classical values into quantum rotational angles. This will eventually introduce heterogeneity into quantum data.

\subsection{Simulation Results on QFL}
\begin{figure*}
    \centering
    \footnotesize
    \begin{subfigure}[t]{0.33\linewidth} 
        \centering
        \includegraphics[width=\linewidth]{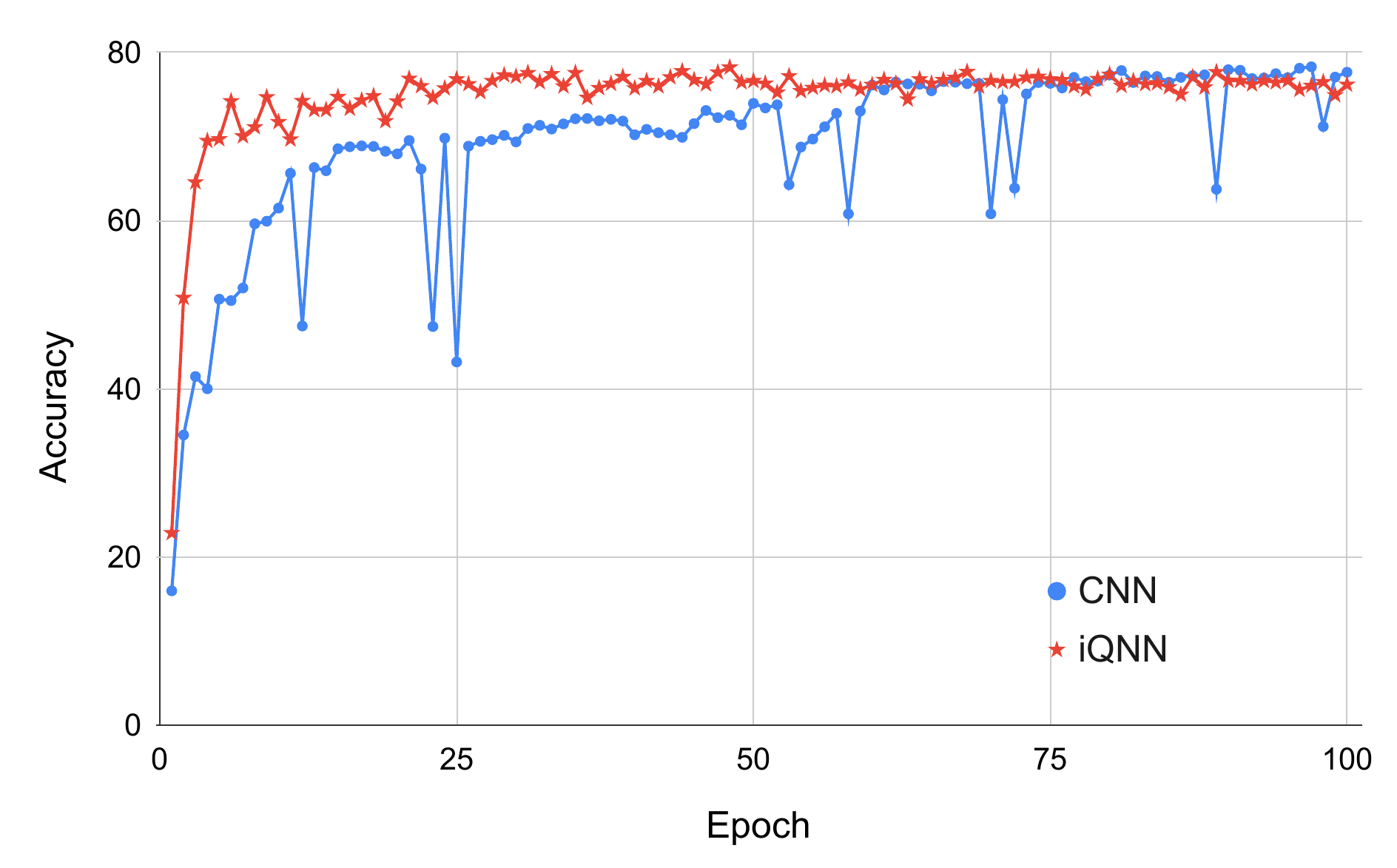}
        \caption{Image data}
        \label{fig: quantum_compa}
    \end{subfigure}
    \hfill 
    \begin{subfigure}[t]{0.33\linewidth} 
        \centering
        \includegraphics[width=\linewidth]{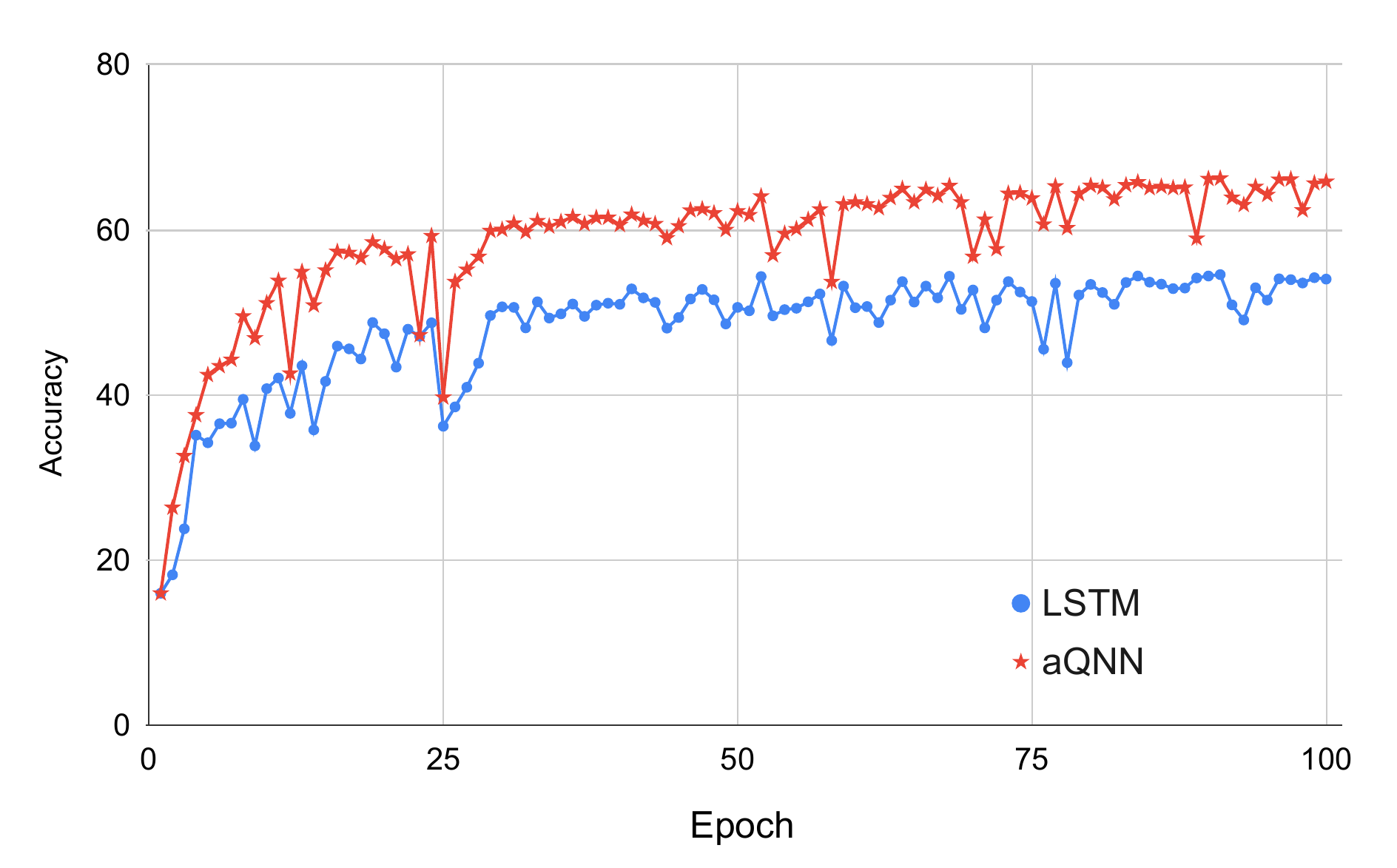}
        \caption{Audio data}
        \label{fig: quantum_compb}
    \end{subfigure}
    \hfill
    \begin{subfigure}[t]{0.33\linewidth} 
        \centering
        \includegraphics[width=\linewidth]{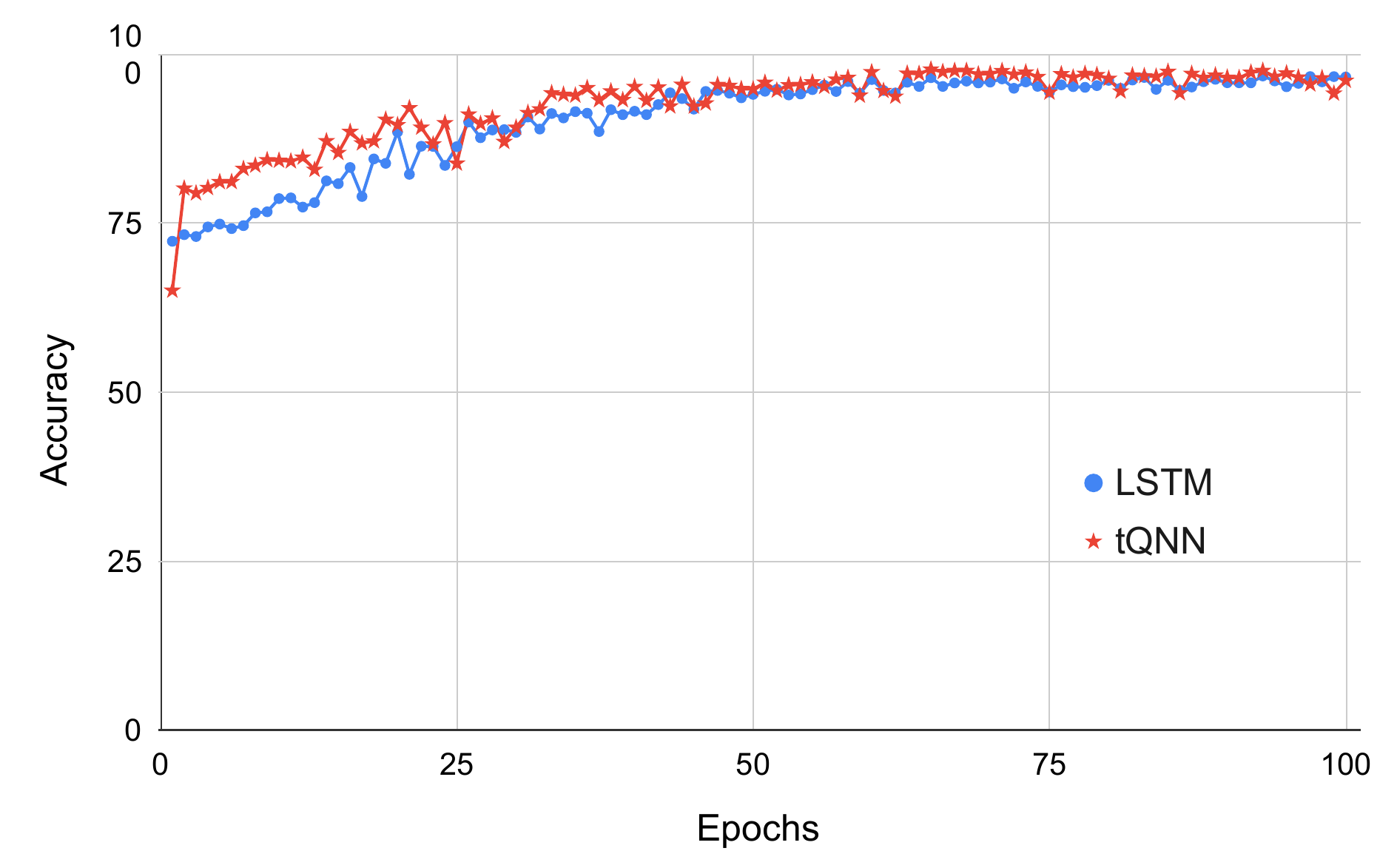}
        \caption{Text data}
        \label{fig: quantum_compc}
    \end{subfigure}
    \vspace{-5pt} 
    \caption{Performance comparison between quantum model approach and classical approach. In (a) for image data classification, we use CNN for the classical approach and the iQNN model for the quantum approach. Similarly, in (b) and (c), we use LSTM for the classical approach and aQNN and tQNN for the quantum approach in audio and text data respectively.}
    \label{fig: quantum_comp}
    \vspace{-3mm}
\end{figure*}

\textbf{Effect on using the quantum model.} This part of our simulation empirically answers a very critical query: \textit{What is the advantage of QFL over classical FL?}. Note that we haven't yet incorporated a multimodal approach in this part of the simulation. Our simulation results (Figure~\ref{fig: quantum_comp}) show improved performance with quantum models. For comparison, CNN is used for image classification and LSTM for audio and text in FL, while iQNN, aQNN, and tQNN are used for the same tasks in QFL. Results indicate that iQNN converges faster than CNN with similar accuracy (Figure~\ref{fig: quantum_compa}). aQNN outperforms LSTM significantly in accuracy and slightly in convergence speed for audio (Figure~\ref{fig: quantum_compb}). For text, tQNN tends to match LSTM's accuracy (Figure~\ref{fig: quantum_compc}). Thus, QNN models have shown overall better performance in emotion detection datasets, and we proceed with these QNN models in our QFL simulations. This claim is further validated in multiple literature works \cite{chen2021federated, araujo2024quantum, chehimi2023foundations}. Such individual QNN models are later used for modality-specific models in multimodal settings.

\textbf{Ablation study.} In our ablation study, we investigate \textit{how different quantum factors affect the performance of our QFL model}, with a particular emphasis on the number of quantum bits (qubits) and quantum layers.  Table~\ref{tab: qubits} summarizes our experiments comparing model performance with qubit configurations $D_q = 2, 3, 5,$ and $10$. All other parameters and training epochs were consistent. Results show that performance improves with more qubits, with $D_q = 10$ yielding the best results. Further increasing qubits exponentially expands the quantum state, so additional simulations use 10 qubits to balance performance and complexity.

\begin{table}
    \centering
    \footnotesize
    \begin{tabular}{c|c|c|c}
        \hline
        \multirow{2}{*}{$D_q$}& Image Data &Audio Data & Text Data \\
         & Accuracy $\uparrow$ & Accuracy $\uparrow$ & Accuracy $\uparrow$ \\
        \hline
        2  & 52.22\% & 45.67\% & 76.10\%\\
        3  & 68.91\% & 55.74\% & 84.55\%\\
        5  & 75.32\% & 61.02\% &  91.18\%\\
        10 & \textbf{78.50\%} & \textbf{64.33\%} & \textbf{96.29\%}\\ 
        \hline
    \end{tabular}
    \vspace{-1mm}
    \caption{Ablation study to investigate the effects of different qubit counts $D_q$ on QFL model performance.  The qubits being compared were 2, 3, 5, and 10.  We limited the number of qubits to 10 to minimize excessive computational demands and resource consumption.}
    \label{tab: qubits}
    \vspace{-3mm}
\end{table}

Next, we examine the impact of quantum layers ($l$) in our QFL model (Table~\ref{tab: qlayer}) using 1, 2, 3, 5, and 10 layers with fixed settings of qubits and training rounds. Unlike qubit variations, performance did not follow a linear trend. The $l = 1$ layer performed best for text and image data, while $l = 3$ was optimal for audio. Thus, we use $l = 1$ for text and image, and $l = 3$ for audio in further tests.

\begin{table}
    \centering
    \footnotesize
    \begin{tabular}{c|c|c|c}
        \hline
        \multirow{2}{*}{$l$}& Image Data &Audio Data & Text Data \\
         & Accuracy $\uparrow$ & Accuracy $\uparrow$ & Accuracy $\uparrow$ \\
        \hline
        1  & \textbf{78.47\%}& 63.98\% & \textbf{96.31\%} \\
        2  & 78.23\% & 62.08\% & 95.60\%\\
        3  & 77.75\% & \textbf{64.41\%}& 95.97\%\\
        5  & 76.93\% & 60.28\% & 96.03\%\\
        10  & 74.20\% & 61.15\% & 94.50\%\\
        \hline
    \end{tabular}
    \caption{Comparing configurations with 1, 2, 3, 5, and 10 layers to determine the impact of quantum layer depth \(l \) in the QFL model.  The study investigates the trade-offs between model expressiveness and training efficiency, arguing that fewer layers may result in underfitting due to limited expressiveness, whereas more layers improve representational capability at the expense of encountering barren plateaus that impede efficient optimization.}
    \label{tab: qlayer}
    \vspace{-3mm}
\end{table}

\textbf{Effects on Number of Clients.} We evaluated QFL performance with varying client numbers in Table~\ref{tab:data_distribution} across IID and three non-IID data distributions. Increasing the number of clients improved accuracy across all modalities. In the IID setting, increasing clients from 3 to 10 improved image, audio, and text accuracy by 2.22\%, 2.26\%, and 1.07\%, respectively. In non-IID settings, accuracy increased by 1.38\% (image), 2.27\% (audio), and 3.81\% (text). The best results were achieved with $N = 10$ clients, with IID showing the highest overall accuracy.

\begin{table}
  \footnotesize
  \centering
  \begin{tabular}{p{1.4cm}|p{0.9cm}|p{1.4cm}|p{1.4cm}|p{1.15cm}}
    \toprule
    Data Distribution & Clients & Image Acc$\uparrow$&Audio Acc$\uparrow$& Text Acc$\uparrow$\\
    \midrule
    \multirow{3}{*}{IID} 
    & $N$ = 3 &76.88\%&62.29\%& 95.25\%\\
    & $N$ = 5 &78.43\%&63.94\%& 95.79\%\\
    & $N$ = 10 &\textbf{79.10\%}&\textbf{64.55\%}& \textbf{96.32\%}\\
    \midrule
    \multirow{3}{*}{Non-IID} 
    & $N$ = 3 &70.11\%&58.76\%& 88.96\%\\
    & $N$ = 5 &70.58\%&59.87\%& 90.28\%\\
    & $N$ = 10 &\textbf{71.49\%}&\textbf{61.03\%}& \textbf{92.77\%}\\
    \bottomrule
  \end{tabular}
    \caption{Different number of clients $N$ on mmQFL environment in IID and non-IID data distribution. We have used 3, 5, and 10 clients for comparison. Experimenting with the number of clients helps evaluate scalability, generalization, and convergence where adding more clients generally results in more robustness.}
  \label{tab:data_distribution}
  \vspace{-3mm}
\end{table}

\textbf{Effects on small data.}
Table~\ref{tab: datasize} presents the classification accuracy of the QFL model across IID and non-IID data distributions with data sizes ($|D_n| = 10\%, 25\%, 50\%,$ and $100\%$). Results show that larger data sizes improve accuracy in all modalities. In the IID setting, increasing data from 10\% to 100\% boosts image, audio, and text accuracy by 25.82\%, 18.64\%, and 17.34\%, respectively. In non-IID settings, accuracy improves by 23.89\% (image), 19.17\% (audio), and 20.74\% (text). Thus, we can conclude that larger datasets enhance model robustness, though non-IID data reduces overall performance.

\begin{table}[ht]
    \centering
    \footnotesize
    \begin{tabular}{p{1.5cm}|c|c|c|c}
        \hline
        \multirow{2}{*}{\shortstack{Data \\ distribution}} & \multirow{2}{*}{$|D_n|$}& Image Data &Audio Data & Text Data \\
         & & Accuracy $\uparrow$ & Accuracy $\uparrow$ & Accuracy $\uparrow$ \\
        \hline
        \multirow{4}{*}{IID} & 10\%  & 53.10\% & 46.98\% & 79.01\%\\
        & 25\%  & 69.34\% & 56.47\% & 85.54\%\\
        & 50\%  & 74.02\% & 63.20\% &  92.91\%\\
        & 100\% & \textbf{78.92\%} & \textbf{65.62\%} & \textbf{96.35\%}\\ 
        \hline
        \multirow{4}{*}{Non-IID} & 10\%  & 50.12\% & 43.64\% & 73.94\%\\
        & 25\%  & 64.24\% & 53.72\% & 83.15\%\\
        & 50\%  & 71.23\% & 59.92\% &  89.77\%\\
        & 100\% & \textbf{74.01\%} & \textbf{62.81\%} & \textbf{94.68\%}\\ 
        \hline
    \end{tabular}
    \caption{Different data size $|D|$ using QFL environment across iid AND various non-IID data distributions. The data size includes 10\%, 25\%, 50\%, and 100\%.}
    \label{tab: datasize}
    \vspace{-3mm}
\end{table}

\subsection{Simulation Results with Multi-modalties}
\textbf{Effects of individual modality models:} 
Table~\ref{tab:mm} compares classification accuracy across image, audio, and text modalities in IID and non-IID data distributions to assess the influence of multimodal learning. In this setting, the quantum fusion layer with entanglement is not considered and the late-fusion approach is applied. The results show that adopting a single-server multimodal method consistently increases accuracy over separate-server models, with slight improvements of 1.40\% (image), 2.51\% (audio), and 1.07\% (text) in the IID scenario, and 1.65\% (image), 1.52\% (audio), and 1.50\% (text) in the Non-IID environment.  Overall, our findings emphasize the benefits of multimodal integration in QFL, particularly in terms of enhancing feature representation and learning efficiency across a variety of data types. The results show that the multimodal setting provides a more robust representation of the data

\begin{table}
\footnotesize
    \centering
    \begin{tabular}{c|c|c|c|c}
    \hline
        \shortstack{Data\\distribution} & Multimodal & \shortstack{Image\\Accuracy} $\uparrow$ & \shortstack{Audio\\Accuracy} $\uparrow$ & \shortstack{Text\\Accuracy} $\uparrow$\\
    \hline
        \multirow{2}{*}{IID}&\shortstack{QFL} & 78.92\%&65.62\%& 96.35\%\\
        \cline{2-5}
        & \shortstack{mmQFL} & \textbf{80.32\%} & \textbf{68.13\%} & \textbf{97.42\%}\\
        \hline
        \multirow{2}{*}{Non-IID}&\shortstack{QFL} & 74.01\%&62.81\%& 94.68\% \\
        \cline{2-5}
        & \shortstack{mmQFL} & \textbf{75.66\%}& \textbf{64.33\%} & \textbf{96.18\%} \\
        \hline
    \end{tabular}
    \caption{Comparison results between without and with multimodal approaches in both IID and Non-IID data distribution. The multimodal model uses one server whereas the other model has a separate server for each data type.}
    \label{tab:mm}
    \vspace{-3mm}
\end{table}
\begin{table}
\footnotesize
    \centering
    \begin{tabular}{|c|c|c|c|c|}
    \hline
        \shortstack{Data \\Distribution}&\shortstack{Missing \\Modality} &Quantity& \shortstack{Without \\MMA} & \shortstack{With \\MMA}\\
        \hline
        \multirow{9}{*}{IID} & None & - & 96.36\% & 96.41\%\\ 
         \cline{2-5}
        & \multirow{3}{*}{Image}& 1\% & 88.21\% & 91.90\% \\
         \cline{3-5}
        & & 10\% &78.89\% & 84.50\%\\
         \cline{3-5}
        & & 20\% & 72.18\%& 80.02\%\\
         \cline{2-5}
        & \multirow{3}{*}{Audio}& 1\% & 90.55\%& 92.26\%\\
         \cline{3-5}
        & & 10\% & 85.73\%& 89.31\%\\
         \cline{3-5}
        & & 20\% & 82.47\%& 86.52\%\\
         \cline{2-5}
        & \multirow{3}{*}{Text}& 1\% & 82.75\% & 87.02\%\\
         \cline{3-5}
        & & 10\% & 73.30\%& 80.15\%\\
         \cline{3-5}
        & & 20\% & 64.64\%& 75.30\%\\
         \hline
        \multirow{9}{*}{Non-IID} & None & - & 82.36\% & 82.41\%\\ 
         \cline{2-5}
        & \multirow{3}{*}{Image}& 1\% & 78.36\%& 80.05\%\\
         \cline{3-5}
        & & 10\% & 72.40\%& 76.98\%\\
         \cline{3-5}
        & & 20\% & 69.22\% & 74.71\% \\
         \cline{2-5}
        & \multirow{3}{*}{Audio}& 1\% & 79.84\%& 81.33\%\\
         \cline{3-5}
        & & 10\% & 75.72\%& 78.50\%\\
         \cline{3-5}
        & & 20\% & 72.01\%& 75.45\%\\
         \cline{2-5}
        & \multirow{3}{*}{Text}& 1\% & 75.44\% & 77.35\%\\
         \cline{3-5}
        & & 10\% & 68.93\%& 74.01\% \\
         \cline{3-5}
        & & 20\% & 63.58\%& 70.30\%\\
         \hline
    \end{tabular}
    \caption{Comparison results between different missing modality levels $m$ in without MMA and with MMA in both IID and non-IID data distribution. The missing levels used in this comparison are 1\%, 10\%, and 20\%, where 1\% indicates 1 in 100 data points is missing.}
    \label{tab: mm}
    \vspace{-3mm}
\end{table}

\begin{figure}
    \centering
    \includegraphics[width=0.99\linewidth]{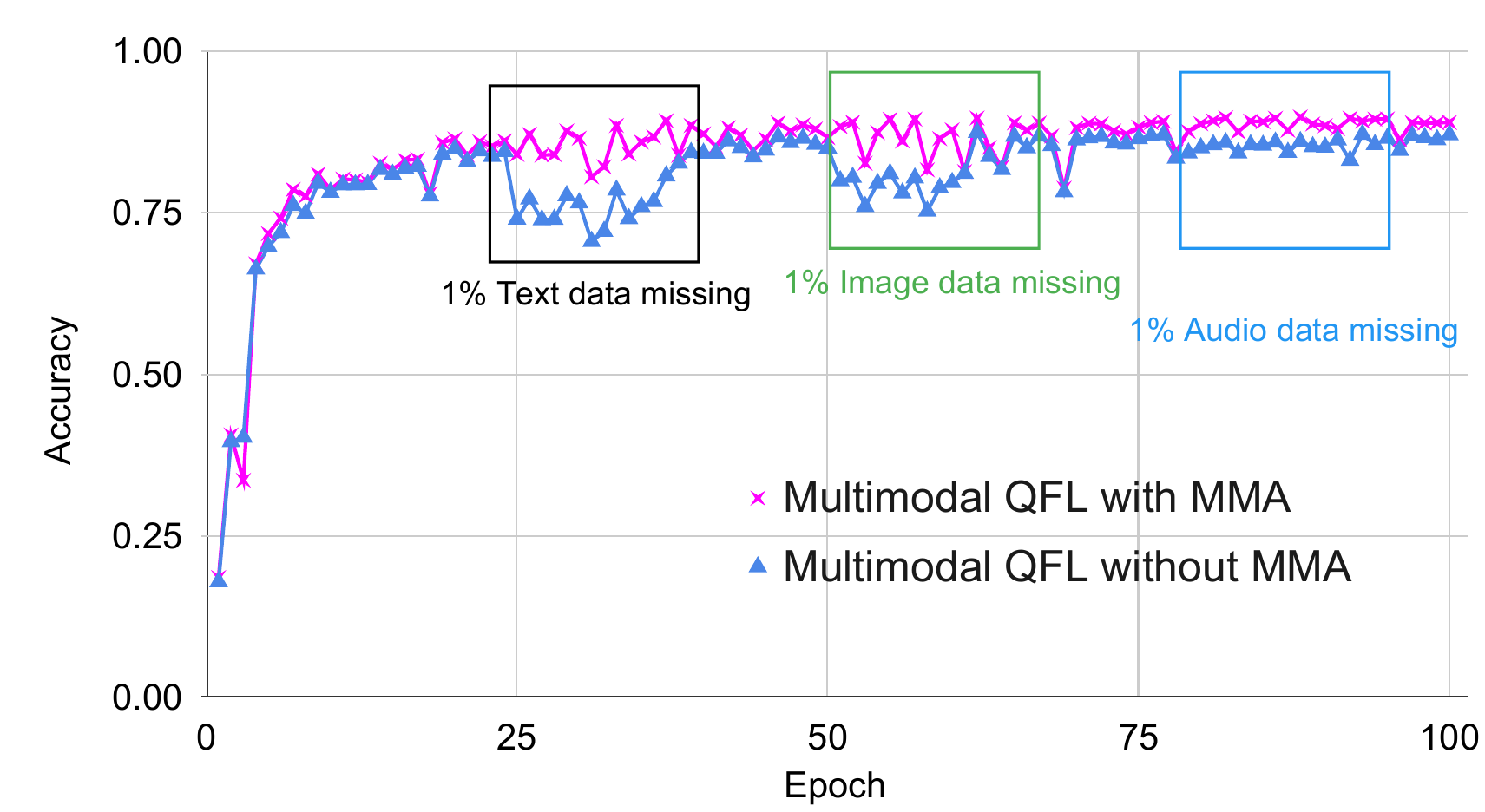}
    \caption{Comparison results between without MMA and with MMA in non-IID data distribution. In epochs 25-35, 50-60, and 80-90, 1\% of audio, image, and text data is missing, respectively.}
    \label{fig: mm}
    \vspace{-3mm}
\end{figure}

\textbf{Simulation Results Missing Modalities:}
Finally, we add a quantum fusion layer before measurement that interconnects the multiple modalities in \textit{a single quantum model} to establish a robust inter-modality relationship for a final prediction as well as a context vector-based approach to tackle the issue of missing modalities. Table~\ref{tab: mm} compares model performance with and without the MMA strategy for different amounts of missing modalities (Image, Audio, Text) at 1\%, 10\%, and 20\%, across IID and non-IID data distributions. Under IID distribution, baseline accuracy without missing modalities is acceptable (~96.\%) but considerably decreases with modality deficits.  For example, eliminating 20\% of the Text modality lowers accuracy from 96.36\% to 64.64\% (without MMA) as the text was the most contributing factor in the overall model's performance.  Introducing MMA significantly improves accuracy in these difficult circumstances; with the same 20\% text modality loss, MMA enhances accuracy from 64.64\% to 75.30\%. Similarly, for a non-IID distribution, baseline accuracy (82.36\%) declines significantly as modality loss increases.  For example, a 20\% missing image modality reduces accuracy from 82.36\% to 69.22\%, with MMA slightly recovering to 74.71\%.  The benefit of MMA is consistent across all modalities and situations, indicating its usefulness in decreasing performance decline when modalities are absent. Figure~\ref{fig: mm} also describes the advantage of our missing modalities agonistic approach where the accuracy graph is sustained even in changing missing modalities scenarios. 

\subsection{Comparison with State-of-the-art Approaches} 
Finally, we compare our method with other state-of-the-art approaches to evaluate the performance of our model. For the comparison, we choose classical unimodal models (\cite{zhang2019face} for image, \cite{noroozi2017audio} for audio, and \cite{kratzwald2018deep} for text), multimodal ML approach Emonets \cite{kahou2016emonets}, multimodal FL approach Fedmultimodal \cite{feng2023fedmultimodal}, multimodal QML approach QNMF \cite{qu2023qnmf}, QFL approach QFSM \cite{qu2024qfsm}, and our proposed mmQFL approach. We compare in both IID and non-IID scenarios. Table \ref{tab: sota} shows the accuracy results for various approaches when 10\% of data from all modalities is absent.  Our mmQFL technique has the highest combined accuracy of 85.61\% in IID settings and 79.27\% in non-IID conditions, exceeding all other approaches.
\begin{table}
    \centering
    \footnotesize
    \begin{tabular}{c|c|c}
        \hline
          \multirow{2}{*}{Method}& IID & Non-IID\\
           & Combined Acc. $\uparrow$ &  Combined Acc. $\uparrow$\\
        \hline
          Unimodal \cite{zhang2019face, noroozi2017audio, kratzwald2018deep}  & 65.19\% & 61.19\%\\
         Emonets \cite{kahou2016emonets}&72.44\% & 67.97\%\\
         Fedmultimodal \cite{feng2023fedmultimodal} & 75.36\%& 69.40\%\\
         QFSM \cite{qu2024qfsm} & 74.23\% & 70.61\%\\
         QNMF \cite{qu2023qnmf} & 78.75\%& 72.02\%\\
         mmQFL & 85.61\%& 79.27\%\\
        \hline
    \end{tabular}
    \caption{Comparison of our approach with other state-of-the-art approaches in both IID and non-IID data distribution with 10\% missing data of all modalities. For comparison, we have selected unimodal approaches for all modalities and other relevant multimodal approaches in emotion detection.}
    \label{tab: sota}
    \vspace{-3mm}
\end{table}

 Unimodal approaches have the lowest accuracy (65.19\% for IID, 61.19\% for non-IID) because they lack cross-modal integration, which is critical in emotion identification.  Multimodal approaches, such as Emonets and Fedmultimodal, exhibit gains (72.44\% and 75.36\% in IID, respectively), although they still fall short of quantum-based approaches.  QNMF, which includes quantum principles, achieves 78.75\% accuracy in IID situations and 72.02\% in non-IID conditions, demonstrating the benefits of quantum-enhanced modeling.

 Our mmQFL method outperforms all other approaches and yields an improved accuracy of 7.25\% by combining quantum federated learning with multimodal processing, proving the benefit of quantum fusion layers, and addressing missing modality awareness.  The results show that our proposed approach not only improves multimodal learning but also provides a strong foundation to manage missing modalities, making it ideal for real-world applications.


\section{Conclusions} 
The paper presented a novel mmQFL framework that improved the QML model performance by incorporating multiple data modalities to generate a single classification result. A quantum fusion layer approach with trainable parameters and entanglement was introduced to establish cross-modal correlations directly within the quantum state space, enhancing the combined feature representations. To mitigate the adverse effects of missing modalities, the MMA mechanism was employed to identify and isolate the respective modality using a context vector approach and no-op gates. Extensive simulations on the CMU-MOSEI dataset demonstrated that the proposed framework achieved superior results compared to existing methods in both IID and non-IID distributions by revealing improved accuracy and robustness against missing modalities.
{
    \small
    \bibliographystyle{ieeenat_fullname}
    \bibliography{main}
}


\end{document}